\documentclass[10pt,conference]{IEEEtran}
\IEEEoverridecommandlockouts
\usepackage{cite}
\usepackage{amsmath,amssymb,amsfonts}
\usepackage{algorithmic}
\usepackage{graphicx}
\usepackage{textcomp}
\usepackage{xcolor}

\usepackage{algorithm}
\usepackage{booktabs}
\usepackage{siunitx} 
\usepackage{array}
\usepackage{multirow}
\usepackage{url}
\usepackage{newfloat}
\usepackage{listings}
\def\BibTeX{{\rm B\kern-.05em{\sc i\kern-.025em b}\kern-.08em
    T\kern-.1667em\lower.7ex\hbox{E}\kern-.125emX}}
\begin{document}

\title{Time-Prompt: Integrated Heterogeneous Prompts for Unlocking LLMs in Time Series Forecasting
\thanks{This work was supported in part by the National Natural Science Foundation of China under Grant 62394341, Grant 62027811. The authors acknowledge the computational resources provided by the High-Performance Computing Centre of Central South University, China.}
}
\author{
	\IEEEauthorblockN{Zesen Wang, Yonggang Li, Lijuan Lan\IEEEauthorrefmark{1}\thanks{Corresponding author: lijuan.lan@csu.edu.cn}}
	\IEEEauthorblockA{Central South University, Changsha, China}
}

\maketitle
\begingroup
\renewcommand{\thefootnote}{}
\footnotetext{\textcopyright~2026 IEEE. Personal use of this material is permitted. Permission from IEEE must be obtained for all other uses, in any current or future media, including reprinting/republishing this material for advertising or promotional purposes, creating new collective works, for resale or redistribution to servers or lists, or reuse of any copyrighted component of this work in other works.}
\endgroup

\begin{abstract}
	Time series forecasting aims to model temporal dependencies among variables for future state inference, holding significant importance and widespread applications in real-world scenarios. Although deep learning-based methods have achieved remarkable progress, they often struggle with long-term forecasting and few-shot scenarios. Recent research demonstrates that large language models (LLMs) achieve promising performance in time series forecasting, but the full potential of LLMs in understanding time series remains largely untapped. To address this, we propose Time-Prompt, a framework for activating LLMs for time series forecasting. Specifically, we first construct a unified prompt paradigm with learnable soft prompts to guide the LLMs' behavior and textualized hard prompts to enhance the time series representations. Second, to enhance LLMs' comprehensive understanding of the forecasting task, we design a semantic space embedding and cross-modal alignment module to facilitate the fusion of temporal and textual data. Finally, we efficiently fine-tune the LLMs' parameters using time series data. Furthermore, we apply our method to carbon emission forecasting, contributing to the technical advancements aiding global carbon neutrality. Comprehensive evaluations on 6 public datasets and 3 carbon emission datasets demonstrate that Time-Prompt is a powerful framework for time series forecasting. Our code is publicly available at \url{https://github.com/SanMuGuo/Time-Prompt}.
\end{abstract}


\section{Introduction}
\par Time series forecasting, which utilizes past observations to predict future data, finds wide applications in domains such as electricity, weather, and carbon emissions \cite{wang2024deep}. Over the past five years, neural networks with diverse architectures—including linear models \cite{zeng2023transformers, wang2024timemixer}, convolutional neural networks \cite{wang2023micn, wu2023timesnet}, recurrent neural networks \cite{gu2023mamba}, and Transformers \cite{vaswani2017attention}—have been applied to time series forecasting. However, existing models face two critical challenges: On one hand, they rely heavily on large-scale labeled samples for parameter optimization, often suffering from limited generalization capability in cross-domain scenarios, which results in performance barriers under few-shot or zero-shot forecasting settings \cite{gruver2023large,liu2024lstprompt}. On the other hand, existing methods struggle to capture long-term temporal dependencies, resulting in suboptimal long-term forecasting performance \cite{chen2024pathformer,ni2025timedistill}.

\par
\begin{figure*}[t] 
	\centering 
	\includegraphics[width=0.95\linewidth]{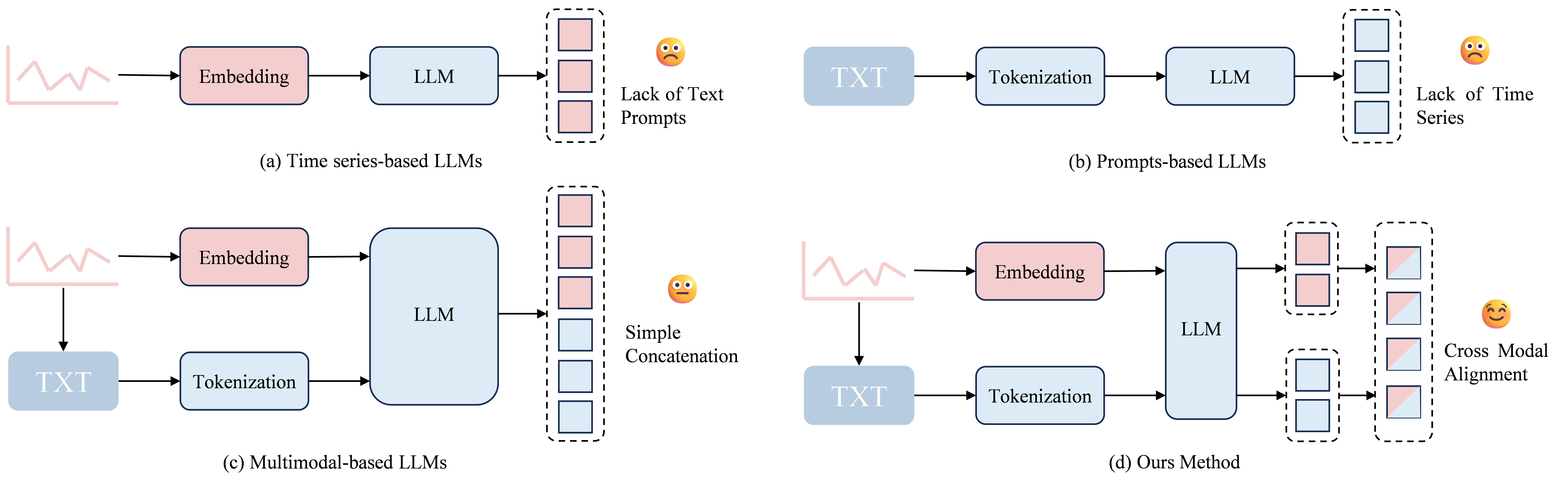}
	\caption{a) Time series-based LLMs: Lack textual information, failing to leverage LLMs' potential for time series forecasting. (b) Prompt-based LLMs: Struggle to capture fine-grained temporal patterns from textual prompts. (c) Multimodal-based LLMs: Time series are processed via an encoder, and text via LLMs, separately. (d) Multimodal-based LLMs: Time series and text information are processed by LLMs in a unified manner.} 
	\label{LLM} 
\end{figure*}

 Pretrained foundation models, such as Large Language Models (LLMs), have demonstrated strong few-shot and zero-shot learning capabilities in computer vision and natural language processing. As pre-trained models advance, research is increasingly split between harnessing LLMs for time series forecasting \cite{shyalika2024time, jin2023large, zhang2024large} and training time series foundation models \cite{goswami2024moment, ansari2024chronos}. Considering the high computational cost of foundation models, our work focuses on the former. As shown in Figure \ref{LLM}, the most common LLM-based approaches are: (a) Time series-based LLMs: Projecting the time series into the LLM’s semantic space via an embedding layer \cite{zhou2023one}; (b) Prompt-based LLMs: converting temporal data into textual modality through prompt engineering \cite{liu2024lstprompt}; 
(c) Multimodal-based LLMs: Processing of time series and text separately \cite{zhou2025balm}; (d) Multimodal-based LLMs: Feeding unified multimodal information into the LLMs \cite{zhao2025enhancing}.

\par Currently, research primarily focuses on two main types (c) and (d). For example, TimeCMA first encodes the multimodal data, where the time series is processed by an encoder and the text by a frozen LLM. Subsequently, cross-modal alignment is performed, and finally, the prediction result is obtained through a decoder \cite{liu2024timecma}. In contrast, Time-LLM processes both time series and text uniformly through a frozen LLM, which then projects the target value \cite{jin2023time}.

\par Despite the state-of-the-art performance of existing methods, Tan et al. \cite{tan2024language} argue that LLMs are ill-suited for time series forecasting. However, Li et al. \cite{li2025tsfm} demonstrate that LLMs hold latent capabilities for this task which currently remain dormant. Aligning with this finding, we believe that through effective guidance strategies, the potential of LLMs in time series forecasting can be fully exploited. This is because, for LLMs, textual reasoning is fundamentally a form of discrete-value reasoning. Viewed from this perspective, it is quite similar to the task of time series forecasting, with the main distinction being that time series consists of continuous values. Furthermore, LLMs have exhibited certain mathematical reasoning abilities, which may suggest a latent capacity for handling continuous values.

\par To unlock the potential of LLMs for time series forecasting, inspired by recent research \cite{liu2024timecma,cao2023tempo,pan2024s,jin2023time}, we propose Time-Prompt, an LLM-empowered framework that activates the potential of LLMs for time series forecasting through time series reprogramming, dual-path prompting, cross-modal alignment, and LoRA. We design a dual-path prompting framework that activates the LLM’s specific patterns with soft prompts and enhances its time series representations with hard prompts. We also reprogram the time series to endow it with a text-like representation.  Furthermore, we construct a cross-modal alignment module to eliminate the modality gap between text and time, and we efficiently fine-tune the LLM with LoRA to enhance its ability to process continuous values.

\par In addition, given that carbon emissions have become a globally critical issue, we conduct a cross-scale study on carbon emission forecasting—spanning global, national, and urban levels—to provide technical support for the strategic goals of global carbon neutrality. Notably, beyond publicly available benchmark datasets, we collect the Munich carbon emission dataset to validate the effectiveness of Time-Prompt in practical scenarios \cite{lan2018self,lan2019vcsel}.
\par The main contributions of this study are summarized as follows:

\begin{itemize}
	\item We propose Time-Prompt, a framework that activates the potential of LLMs in time series forecasting through strategies such as dual-path prompting, cross-modal alignment, and efficient fine-tuning.
	\item We design a dual-path prompting strategy, with learnable soft prompts guiding the LLM’s behavior and hard prompts extracting enhanced time series representations.
	\item Unlike prior methods that simply concatenate multimodal tokens, we construct a semantic space embedding module and a cross-modal alignment module to enable cross-modal fusion of temporal data and textual information, eliminating the modality gap.
	\item Extensive experiments on 6 benchmark datasets and 3 carbon emission datasets demonstrate that Time-Prompt achieves superior performance.
\end{itemize}

\section{Related Work}
\par Deep learning has demonstrated superior performance in time series forecasting, primarily categorized into MLP-based, CNN-based, RNN-based, and Transformer-based architectures. Transformer-based models have emerged as the core research direction in recent years by capturing long-term dependencies and inter-variable correlations through self-attention mechanisms, achieving better predictive performance compared to other architectures. iTransformer treats multivariate time series as independent variable sequences, capturing cross-variable dependencies via attention mechanisms \cite{liu2023itransformer}. PatchTST divides time series into fixed-length patches, enhancing local pattern recognition through patch-level tokenization \cite{Yuqietal-2023-PatchTST}. Furthermore, addressing the challenge of non-stationary time series, Non-stationary Transformer proposes series stationarization and de-stationary attention to mitigate non-stationarity in raw time series \cite{liu2022non}. Other variants include Autoformer \cite{wu2021autoformer}, which integrates decomposition frameworks with autocorrelation mechanisms, and Fedformer \cite{zhou2022fedformer}, which employs frequency-domain enhanced decomposition strategies, both significantly improving long-term forecasting accuracy.
\begin{figure*}[t] 
	\centering 
	\includegraphics[width=0.95\linewidth]{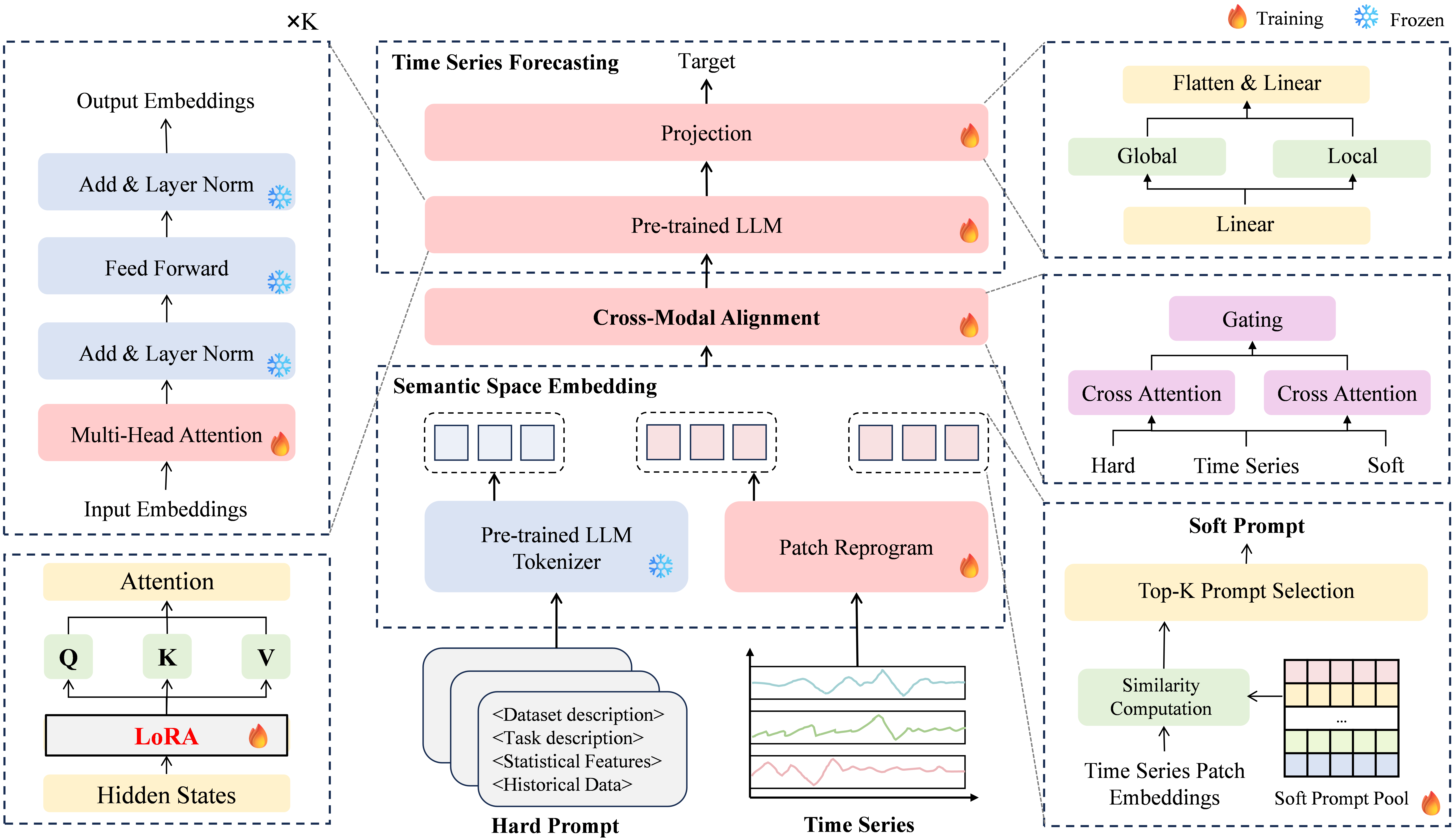}
	\caption{Time-Prompt framework. First, hard prompts and soft prompts are constructed based on time series. Second, the time series and prompts are embedded into the LLM's semantic space, followed by cross-modal alignment. Finally, the fused information is fed into the pretrained LLM and projected through output layers to generate forecasting results.} 
	\label{framework} 
\end{figure*}
\par Beyond Transformers, DLinear \cite{zeng2023transformers} and TimeMixer \cite{wang2024timemixer} show that simple linear or multi-scale mixing designs remain competitive. CNN-based methods such as SCINet \cite{liu2022scinet} and WaveNet \cite{van2016wavenet} leverage receptive fields to capture temporal patterns, while Mamba \cite{gu2023mamba} reestablishes state-space models as an efficient paradigm for long-term dependency modeling. However, these deep learning frameworks remain constrained by sample size, exhibiting suboptimal performance in few-shot or zero-shot scenarios.
\par Recently, leveraging the cross-domain transfer capabilities of LLMs, they have demonstrated exceptional performance in time series forecasting tasks, significantly outperforming traditional methods, especially in few-shot scenarios \cite{liu2024autotimes}. As shown in Figure \ref{LLM}, the most prevalent current approaches include: (a) Time series-based LLMs; (b) Prompt-based LLMs; (c) Multimodal-based LLMs; (d) Multimodal-based LLMs. Time series–based LLMs project time series into the semantic space of LLMs via an embedding layer, and subsequently map the embedding vectors to obtain prediction values \cite{zhou2023one}. In contrast, Prompt-based LLMs construct the time series as a text modality \cite{xue2023promptcast}. Lstprompt is a typical Prompt-based method that proposes a prompt construction technique, guiding the LLM to make predictions via a Chain-of-Thought approach \cite{liu2024lstprompt}. 
\par However, single-modality time series forecasting often suffers from limited representational capacity, leading to suboptimal prediction performance. Therefore, many studies attempt to integrate Type (a) and Type (b) approaches, leveraging multimodal information to overcome the inherent limitations of single-modality methods. Early efforts predominantly fall into Type (d), which uniformly feeds time series and textual inputs into LLMs to obtain improved feature representations. GPT4MTS, a foundational work in this category, embeds time series and simple text prompts into the semantic space, from which the large model derives a joint representation \cite{jia2024gpt4mts}. Many subsequent works build upon this framework. Time-LLM enhances model comprehension of time series by reprogramming the input time series \cite{jin2023time}. Meanwhile, S²IP-LLM and TEMPO employ learnable soft prompts to mark specific temporal patterns, thereby more effectively activating the large model \cite{cao2023tempo,pan2024s}.
\par Subsequently, given the limitations of LLMs in directly representing time series, researchers have shifted towards a dual-path paradigm (Type (c)), where time series are handled by dedicated encoders and text prompts are input into the LLMs. TimeCMA encodes the time series with an encoder and the text with an LLM, followed by cross-modal alignment, and finally generates the prediction via a unified decoder \cite{liu2024timecma}. In a related approach, another study further enhances the textual information using reinforcement learning \cite{su2025text}.
\par  The problem with Type (c) is that the LLM only plays an auxiliary role, contributing little to time series forecasting. The issue with Type (d), however, is the LLM’s inability to handle continuous values like those in time series. Nevertheless, to bridge the knowledge gaps in these two categories, relevant research targeting both Type (c) \cite{zhou2025balm,su2025text} and Type (d) \cite{chen2025cc,zhang2024dualtime} continues to emerge. Our method falls into Type (d), aiming to unlock the LLM's latent reasoning ability for continuous-valued forecasting through appropriate guidance.

\par \noindent \textbf{Differences from Existing Methods.} Although Time-Prompt builds on established techniques, its novelty lies in the systematic integration of complementary components. In contrast to Time-LLM \cite{jin2023time} and TimeCMA \cite{liu2024timecma} that rely solely on hard prompts, or S$^2$IP-LLM \cite{pan2024s} and TEMPO \cite{cao2023tempo} that use only soft prompts, we unify both in a dual-path framework where soft prompts guide the LLM's behavioral patterns while hard prompts enrich statistical representations. Existing methods simply concatenate prompts with time-series tokens, which causes feature redundancy; we instead introduce a cross-modal alignment module with learnable gating to adaptively fuse information from multiple sources. Finally, unlike prior works that keep the LLM entirely frozen, we apply LoRA to the attention layers, enabling efficient adaptation to continuous-valued signals---one of the most impactful components in our ablation study.

\section{Methodology}

\subsection{Framework Overview}
\par The overall framework of Time-Prompt is illustrated in Figure \ref{framework}. First, we construct a dual-path prompting framework for the time series forecasting task. This framework incorporates both hard prompts (including temporal statistical features and historical data) and learnable soft prompts. Our goal is to guide the LLM’s internal behavioral patterns via soft prompts and enhance time series feature representation through hard prompts. Second, we embed both the time series and text into the LLM’s semantic space via a semantic space embedding module to facilitate better comprehension by the LLM. Subsequently, to address the modality gap between text and time series, we employ cross-modal alignment. Finally, the global and local features of the time series are extracted and projected to obtain the predicted values. Here, the LLM is fine-tuned using LoRA to adapt it from discrete-value reasoning to continuous-value prediction.

\subsection{Dual-Path Prompting}
\par \noindent  \textbf{Soft Prompts.} Soft prompts include prompt pool construction, similarity computation, and Top-k prompt selection. The core idea is to dynamically select soft prompt vectors most relevant to the time series. First, initialize a prompt pool. Then, compute the similarity between the time series data and prompt keys in the pool, selecting the Top-k most relevant prompts. Notably, the prompt pool is adaptively updated during iterations. 

\par \textbf{Prompt Pool Construction.} Given prompt keys $K \in {\mathbb{R}^{P \times D}}$ and prompt pool $V \in {\mathbb{R}^{P \times L \times D}}$, where \textit{P} is the pool size, \textit{L} is the soft prompt length, and \textit{D} is the embedding dimension. The prompt pool can be initialized through multiple strategies: zero initialization, uniform distribution initialization, or initialized from the vocabulary of pretrained LLMs. Notably, the prompt pool is updated during backpropagation. 

\par \textbf{Similarity Computation.} Given the patch-embedded time series $\hat{X} \in {\mathbb{R}^{B \times N \times M \times D}}$, where $B$ is the batch size, $N$ is the number of variables, $M$ is the number of patches, and $D$ is the embedding dimension. To obtain a holistic representation for each variable, we aggregate $\hat{X}$ along the patch dimension via mean pooling: $\bar{X} = \text{MeanPool}(\hat{X}) \in {\mathbb{R}^{B \times N \times D}}$. The similarity between the aggregated representation $\bar{X}$ and the prompt keys $K$ is then computed via dot product:

\begin{equation}
	S = \bar{X}{K^T} \in {\mathbb{R}^{B \times N \times P}}
\end{equation}

\par \textbf{Top-k Prompt Selection.} Based on the similarity matrix $S$, we select the indices of the Top-k prompt keys with the highest similarity scores along the pool dimension for each variable:

\begin{equation}
	I = \mathop{\mathrm{Top\text{-}k}}\limits_{j \in \{1,...,P\}}(S_{:,:,j}) \in \mathbb{Z}^{B \times N \times k}
\end{equation}

\noindent The selected indices $I$ are then used to retrieve the corresponding prompt vectors from the prompt pool $V$, yielding the soft prompts:

\begin{equation}
	\mathrm{softPrompt} = V[I] \in {\mathbb{R}^{B \times N \times k \times L \times D}}
\end{equation}
\begin{figure}[t]
	\centering 
	\includegraphics[width=0.95\linewidth]{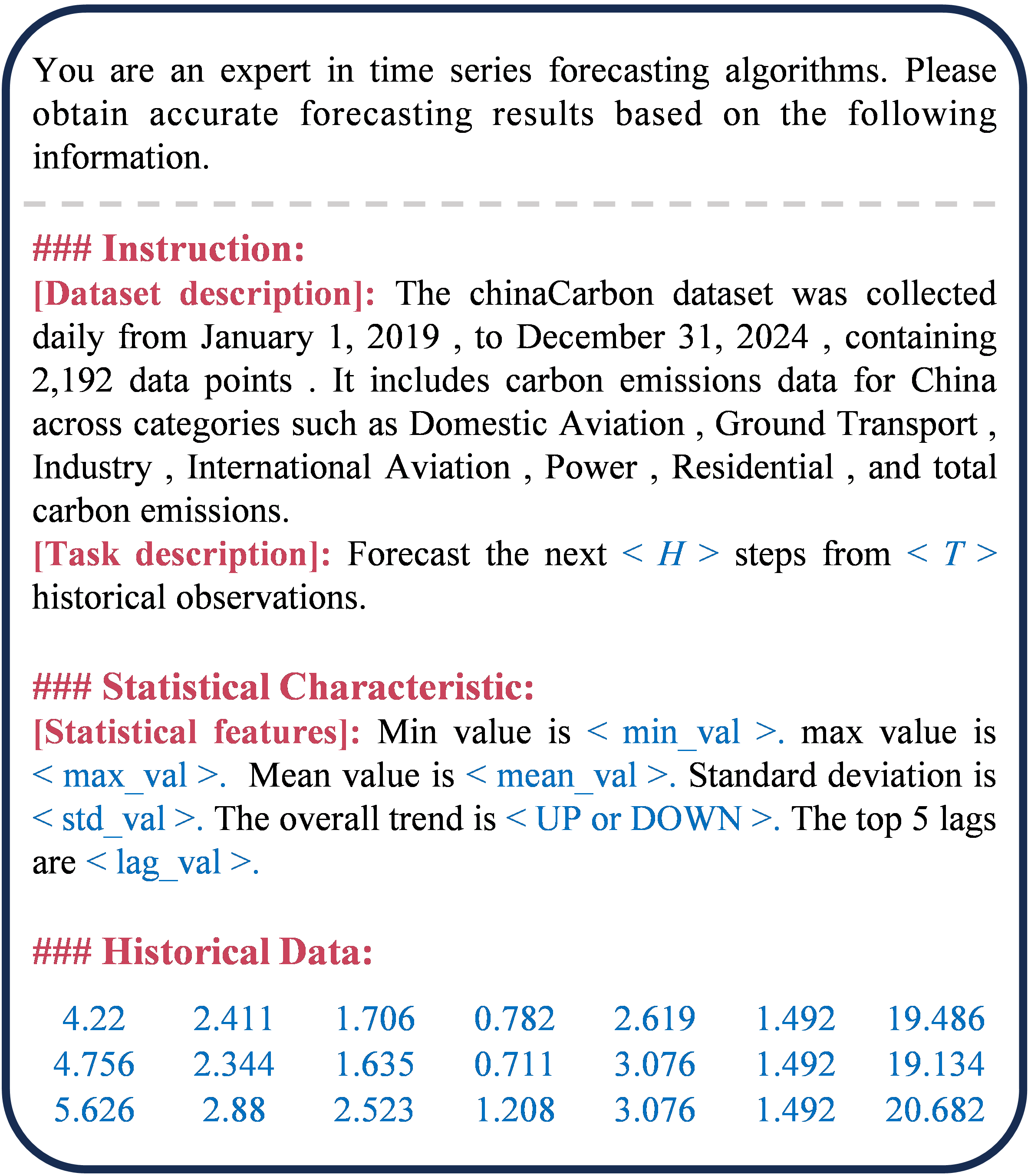}
	\caption{Hard prompt example on chinaCarbon dataset.} 
	\label{hard prompt} 
\end{figure}

\par \noindent  \textbf{Hard Prompts.} Hard prompts provide a direct and effective way to activate the LLM’s time series forecasting capability, as no modality gap exists between the LLM and the prompts. Hard prompts can also be viewed as enhanced representations of the time series. As illustrated in Figure \ref{hard prompt}, our paradigm comprises three key components: (1) Instruction, (2) Statistical Characteristics, and (3) Historical Data. The Instruction provides the LLM with the task context, the statistical characteristics provide informative priors that help the LLM recognize temporal patterns, and the Historical Data is the direct textualization of the time series data. 
\par Notably, all statistical features in the hard prompts, including min, max, mean, standard deviation, median, trend, and top-k lags, are computed exclusively from the input look-back window of each individual sample. No information from the forecasting horizon or test set is used, thereby ensuring strict prevention of data leakage. Furthermore, the data normalization scaler is fitted solely on the training split.

\begin{table*}[t]
	\centering
	\setlength{\tabcolsep}{2pt}
	\scriptsize
	\renewcommand{\arraystretch}{0.8}
	\caption{Long-term forecasting results. {\textbf{Bold}}: the best, {\underline{Underline}}: the second best.}
	\label{long-term}
	\begin{tabular}{@{}c|*{9}{S[table-format=1.3]S[table-format=1.3]|}S[table-format=1.3]S[table-format=1.3]@{}}
		\toprule 
		Methods      & \multicolumn{2}{c|}{Time-Prompt} & \multicolumn{2}{c|}{Time-LLM} & \multicolumn{2}{c|}{TimeCMA} & \multicolumn{2}{c|}{Mamba} & \multicolumn{2}{c|}{TimesNet} & \multicolumn{2}{c|}{DLinear} & \multicolumn{2}{c|}{Autoformer} & \multicolumn{2}{c|}{Crossformer} & \multicolumn{2}{c|}{iTransformer} & \multicolumn{2}{c}{PatchTST} \\
		\midrule
		Metric         & {MSE} & {MAE} & {MSE} & {MAE} & {MSE} & {MAE} & {MSE} & {MAE} & {MSE} & {MAE} & {MSE} & {MAE} & {MSE} & {MAE} & {MSE} & {MAE} & {MSE} & {MAE} & {MSE} & {MAE} \\
		\midrule 
		ETTh1         & \textbf{0.430}&  \textbf{0.439}&  0.449&  \underline{0.445}&  0.449&  0.447&  0.520&  0.492&  0.460&           0.455&0.460&0.457&0.477&0.475&0.549&0.530&0.464&0.455&\underline{0.446}&0.447 \\
		\midrule
		ETTh2         & \textbf{0.368}&  \textbf{0.397}&  \underline{0.380}& \underline{0.403}&  0.399&  0.416&  0.435&  0.447&  0.408&           0.421&0.564&0.519&0.446&0.457&1.920&1.073&0.383&0.406&0.382&0.411 \\
		\midrule
		ETTm1         & \textbf{0.380}&  \textbf{0.398}&  0.421&  0.413&  0.400&  0.410&  0.466&  0.450&  0.408&           0.416&0.404&0.408&0.53&0.496&0.697&0.605&0.407&0.412&\underline{0.390}&\underline{0.405} \\
		\midrule
		ETTm2         & \textbf{0.281}&  \textbf{0.326}&  0.291&  0.335&  0.294&  0.336&  0.331&  0.365&  0.299& 
		\underline{0.333}&0.354&0.401&0.326&0.364&1.725&0.813&0.291&0.334&\underline{0.293}&0.334 \\
		\midrule
		Exchange      & \textbf{0.331}& \textbf{0.390}&  0.361&  \underline{0.403}&  0.448&  0.451&  0.761&  0.584&  0.425&           0.447&\underline{0.340}&0.414&0.843&0.596&0.757&0.645&0.364&0.405&0.391&0.417 \\
		\midrule
		Weather       & \textbf{0.252}& \textbf{0.277}&  0.266&  0.285&  0.257&  0.284&  0.288&  0.314&  0.258&           0.285&0.265&0.317&0.337&0.379&0.259&0.320&0.261&0.281&\underline{0.255}&\underline{0.278}\\
		\midrule
		munichCarbon  & 0.411&  0.464&  0.411&  0.465&  0.433&  0.486&  0.485&  0.518&  
		0.415& 0.469&\underline{0.394}&\underline{0.456}&0.611&0.599&\textbf{0.358}&\textbf{0.437}&0.412&0.465&0.430&0.477 \\
		\midrule
		chinaCarbon   & \textbf{0.571}& \textbf{0.532}& 0.627& 0.557& 0.659& 0.586& 0.660& 0.592& 0.665& 0.588& 0.689& 0.642& 0.642& 0.608& 1.177& 0.834& 0.667& 0.579& \underline{0.604}&\underline{0.546} \\
		\midrule
		worldCarbon   & 0.617&  0.544&  0.649&  0.565& \textbf{0.546}& \textbf{0.518}&  0.687&  0.614&  0.751&           0.639&0.754&0.698&0.653&0.603&0.619&\underline{0.539}&0.735&0.616&\underline{0.615}&0.547 \\
		\bottomrule
	\end{tabular}
\end{table*}

\begin{table*}[t]
	\centering
	\setlength{\tabcolsep}{2pt}
	\scriptsize
	\renewcommand{\arraystretch}{0.8}
	\caption{Few-shot learning on 10\% training data. {\textbf{Bold}}: the best, {\underline{Underline}}: the second best.}
	\label{few-shot}

	\begin{tabular}{@{}c|*{9}{S[table-format=1.3]S[table-format=1.3]|}S[table-format=1.3]S[table-format=1.3]@{}}
		\toprule
		Methods      & \multicolumn{2}{c|}{Time-Prompt} & \multicolumn{2}{c|}{Time-LLM} & \multicolumn{2}{c|}{TimeCMA} & \multicolumn{2}{c|}{Mamba} & \multicolumn{2}{c|}{TimesNet} & \multicolumn{2}{c|}{DLinear} & \multicolumn{2}{c|}{Autoformer} & \multicolumn{2}{c|}{Crossformer} & \multicolumn{2}{c|}{iTransformer} & \multicolumn{2}{c}{PatchTST} \\
		\midrule
		Metric       & {MSE} & {MAE} & {MSE} & {MAE} & {MSE} & {MAE} & {MSE} & {MAE} & {MSE} & {MAE} & {MSE} & {MAE} & {MSE} & {MAE} & {MSE} & {MAE} & {MSE} & {MAE} & {MSE} & {MAE} \\
		\midrule
		ETTh1        & \textbf{0.452}& \textbf{0.445}& \underline{0.477}& 0.460& 0.549& 0.504& 0.716& 0.572& 0.847& 0.632& 0.632& 0.548& 0.510& 0.499& 0.583& 0.531& 0.701& 0.570& 0.477& \underline{0.455} \\
		\midrule
		ETTh2        & \textbf{0.381}& \textbf{0.404}& 0.396& 0.414& 0.416& 0.429& 0.427& 0.448& 0.465& 0.459& 0.435& 0.458& 0.427& 0.446& 0.800& 0.642& 0.445& 0.447& \underline{0.391}& \underline{0.412} \\
		\midrule
		ETTm1        & \textbf{0.394}& \textbf{0.402}& 0.424& 0.411& 0.425& 0.427& 0.511& 0.467& 0.474& 0.454& 0.496& 0.467& 0.623& 0.532& 0.427& 0.439& 0.491& 0.456& \underline{0.414}& \underline{0.406} \\
		\midrule
		ETTm2        & \textbf{0.284}& \textbf{0.327}& 0.297& 0.339& 0.302& 0.342& 0.318& 0.359& 0.311& 0.352& 0.382& 0.431& 0.311& 0.355& 0.592& 0.527& 0.304& 0.345& \underline{0.290}& \underline{0.333} \\
		\midrule
		Exchange     & \textbf{0.337}& \textbf{0.395}& 0.368& \underline{0.407}& 0.423& 0.448& 0.677& 0.554& 0.439& 0.459& \underline{0.368}& 0.431& 0.471& 0.483& 2.043& 1.136& 0.411& 0.442& 0.373& 0.411 \\
		\midrule
		Weather      & \underline{0.265}& \underline{0.286}& 0.269& 0.288& \textbf{0.255}& \textbf{0.282}& 0.293& 0.322& 0.276& 0.301& 0.275& 0.339& 0.348& 0.380& 0.343& 0.397& 0.281& 0.295& 0.273& 0.289 \\
		\midrule
		munichCarbon & \underline{0.406}& \textbf{0.459}& 0.426& 0.477& 0.434& 0.486& 0.513& 0.538& 0.445& 0.491& 0.413& 0.475& 0.605& 0.609& \textbf{0.402}& \underline{0.474}& 0.443& 0.491& 0.423& 0.474 \\
		\midrule
		chinaCarbon  & \textbf{0.653}& \textbf{0.575}& 0.714& 0.606& 0.690& 0.596& 0.785& 0.651& 0.821& 0.655& 0.742& 0.632& 0.766& 0.652& 1.043& 0.833& 0.792& 0.645& \underline{0.665}& \underline{0.580} \\
		\midrule
		worldCarbon  & \textbf{0.705}& \textbf{0.601}& 0.802& 0.669& \underline{0.720}& \underline{0.607}& 0.950& 0.733& 0.959& 0.740& 0.845& 0.699& 0.781& 0.659& 0.966& 0.782& 0.937& 0.727& 0.725& 0.609 \\
		\bottomrule
	\end{tabular}
\end{table*}

\subsection{Cross-Modal Data Fusion}
\par \noindent \textbf{Semantic Space Embedding.} This component primarily involves embedding the time series and prompts into the LLM’s semantic space to enhance the LLM’s understanding of the variables. For soft prompts, they are generated directly in the embedding space and thus require no embedding.

\par \textbf{Time Series.} First, segment the time series into patches, then project them into the LLM's embedding space to obtain $\hat X \in {\mathbb{R}^{B \times N \times M \times D}}$, where \textit{M} is the number of patches and $D$ is the embedding dimension. Given the pre-trained LLM's vocabulary $E \in {\mathbb{R}^{V \times D}}$, where $V$ is the vocabulary size, we realign time series data with the vocabulary through reprogramming, achieving textual representation of temporal data to enhance LLM comprehension. Specifically, learnable projection matrices generate query matrices $Q_r = \hat X{W^Q}$, key matrices $K_r = E{W^K}$, and value matrices $V_r = E{W^V}$  for each attention head. Then, multi-head cross-attention produces the reprogramming temporal representations. 

\begin{equation}
	\tilde X = {\rm{MHCA}}(Q_r,K_r,V_r) \in {\mathbb{R}^{B \times N \times M \times D}}
\end{equation}

\par \textbf{Hard Prompt.} Given the hard prompt $T$, map it to discrete tokens using the LLM's pretrained tokenizer. Then, convert these discrete tokens into continuous vector representations via the LLM's input embedding layer. During this process, the LLM's tokenizer remains frozen with no parameter updates during training. 

\begin{equation}
	{\rm{hardPrompt}} = {\rm{Embed}}({\rm{Tokenize}}(T)) \in {\mathbb{R}^{B \times N \times L \times D}}
\end{equation}

\par \noindent  \textbf{Cross-Modal Alignment.} This module is designed to achieve cross-modal fusion among the time series data, soft prompts, and hard prompts to eliminate the modality gap. Many existing works simply concatenate these variables, which can lead to severe feature redundancy when the number of prompt tokens is large \cite{jin2023time}. Our specific approach is as follows: (a) We use a multi-head cross-attention mechanism to align the soft prompt with the time series, obtaining a fused representation $Z_1$.  Similarly, the hard prompt is aligned with the time series to obtain $Z_2$. (b) We then fuse the original time series and the fused representations $Z_1$ and $Z_2$ via a gating mechanism, yielding the final representation $Z$.

\begin{equation}
	Z = g_0 \odot \tilde{X} + g_1 \odot Z_1 + g_2 \odot Z_2
\end{equation}

\noindent where $g_0, g_1, g_2 \in \mathbb{R}$ are learnable scalar gating weights satisfying $g_0 + g_1 + g_2 = 1$ via softmax normalization.

\begin{table}[t]
	\centering
	\caption{Ablation Experiment. SP: Soft Prompt, HP: Hard Prompt, CMA: Cross-Modal Alignment. \textbf{Bold}: the best.}
	\label{ablation}
	\setlength{\tabcolsep}{3pt}
	\scriptsize
	\resizebox{\linewidth}{!}{%
	\begin{tabular}{@{}l l c c c c c@{}}
		\toprule
		Dataset & Metric & Time-Prompt & w/o SP & w/o HP & w/o CMA & w/o LoRA \\
		\midrule
		\multirow{2}{*}{ETTh1}
		& MSE & \textbf{0.430} & 0.434 & 0.437 & 0.434 & 0.435 \\
		& MAE & \textbf{0.439} & 0.441 & 0.442 & 0.442 & 0.440 \\
		\midrule
		\multirow{2}{*}{ETTm1}
		& MSE & \textbf{0.380} & 0.383 & 0.384 & 0.386 & 0.386 \\
		& MAE & \textbf{0.398} & 0.399 & 0.400 & 0.402 & 0.401 \\
		\midrule
		\multirow{2}{*}{munichCarbon}
		& MSE & \textbf{0.411} & 0.413 & 0.414 & 0.414 & 0.416 \\
		& MAE & \textbf{0.464} & 0.466 & 0.467 & 0.467 & 0.468 \\
		\midrule
		\multirow{2}{*}{chinaCarbon}
		& MSE & \textbf{0.571} & 0.581 & 0.584 & 0.587 & 0.583 \\
		& MAE & \textbf{0.532} & 0.540 & 0.540 & 0.544 & 0.538 \\
		\midrule
		\multirow{2}{*}{worldCarbon}
		& MSE & \textbf{0.617} & 0.636 & 0.633 & 0.633 & 0.630 \\
		& MAE & \textbf{0.544} & 0.558 & 0.556 & 0.556 & 0.552 \\
		\bottomrule
	\end{tabular}%
	}
\end{table}

\subsection{Time Series Forecasting}
\par Many existing works rely solely on frozen LLMs. Given the high dimensionality of the LLM’s semantic space, they often resort to direct truncation to reduce the computational overhead of the high-dimensional semantic space. This approach, however, is detrimental to activating the LLM’s potential for time series forecasting. In this module, we first employ LoRA to efficiently fine-tune the LLM, enhancing its ability to comprehend continuous values. Second, to address the excessively high semantic dimensionality, we project the LLM’s output representations into a lower-dimensional space to obtain local features, aggregate them via mean pooling to capture global patterns, and then fuse local and global features through a linear projection. The fused representation is passed to an output head that directly produces the final prediction $Y$ in a single forward pass, without any autoregressive decoding.

\par Regarding the training strategy, the pre-trained GPT-2 backbone is kept frozen, with only the LoRA adapters (applied to the attention layers, rank $r=8$) receiving gradient updates. The tokenizer and input embedding layer of the LLM also remain frozen throughout training. All other modules are fully trainable, including the patch embedding, reprogramming layer, soft prompt pool, cross-modal alignment, gating fusion weights, feature projection, and output head. This selective fine-tuning strategy preserves the LLM’s pre-trained linguistic knowledge while efficiently adapting it to time series through lightweight parameter updates.

\section{Experiments}
\par  \textbf{Dataset.} We utilize six widely-used public datasets covering multiple domains, including electricity, finance, and weather, as well as three carbon emission datasets with different spatial scales. We conduct extensive experiments across these nine datasets to demonstrate the superiority of Time-Prompt, including tasks such as long-term forecasting and few-shot forecasting. Notably, we introduce a real-world dataset (munichCarbon)—distinct from publicly available benchmarks—to validate the model's effectiveness in practical scenarios.
\par \textbf{Baseline.} We select 9 representative SOTAs from 5 categories: LLM-based: Time-LLM \cite{jin2023time}, TimeCMA \cite{liu2024timecma}; Transformer-based: Autoformer \cite{wu2021autoformer}, Crossformer \cite{zhang2023crossformer}, iTransformer\cite{liu2023itransformer}, PatchTST \cite{Yuqietal-2023-PatchTST}; CNN-based: TimesNet \cite{wu2023timesnet}; RNN-based: Mamba\cite{gu2023mamba}; Linear-based: DLinear \cite{zeng2023transformers}.
\par \textbf{Implementation. }We use GPT-2 as the backbone LLM \cite{radford2019language}. LoRA is applied to the attention projection layers with rank $r=8$, scaling factor $\alpha=16$, and dropout rate 0.1. The optimizer is Adam with an initial learning rate of $1 \times 10^{-4}$, scheduled by OneCycleLR. The batch size is 32, training runs for 10 epochs with early stopping patience of 3. The random seeds are fixed at 2021, 2023, and 2025 for reproducibility. All experiments are independently repeated under these three seeds; Tables~\ref{long-term}, \ref{few-shot}, and \ref{ablation} report the average performance across seeds and across all four forecasting horizons. Tables~\ref{analysis} and \ref{stability} report results for a single representative horizon ($H=96$) to provide fine-grained analysis.
\subsection{Long-term Forecasting}
\par  \textbf{Setups.} We conduct experiments on 9 datasets, including 6 classical benchmarks (ETTh1, ETTh2, ETTm1, ETTm2, exchange-rate, weather) and 3 carbon emission datasets (munichCarbon, chinaCarbon, worldCarbon). The input sequence length $T$ is fixed at 96, and we adopt four different forecasting horizons: $H \in \{ 60, 90, 120, 150\} $ for chinaCarbon and worldCarbon, and $H \in \{ 96, 192, 336, 720\} $ for the others.
\par  \textbf{Results.} The results summarized in Table \ref{long-term}, averaged over different forecasting horizons, demonstrate that Time-Prompt outperforms all baselines in most cases. Time-Prompt achieves the best performance in 14 out of 18 benchmark settings. Overall, LLM-based methods demonstrate superior performance compared to conventional deep learning approaches. Specifically, Time-Prompt reduces MSE and MAE by 5.5\% and 2.7\% over Time-LLM, by 6.3\% and 4.2\% over TimeCMA, and by 4.3\% and 2.4\% compared to PatchTST—the strongest deep learning baseline.

\subsection{Few-shot Forecasting}
\par  \textbf{Setups.} LLMs demonstrate remarkable few-shot learning capabilities in CV and NLP. We further investigate whether LLMs exhibit similar capabilities in forecasting tasks. Most settings follow the long-term forecasting configuration. This section evaluates model performance under the scenario with only 10\% of training samples.
\par \textbf{Results.} The results summarized in Table \ref{few-shot}, averaged over different forecasting horizons, demonstrate that LLM-based methods significantly outperform deep learning models, confirming that LLMs retain few-shot learning capabilities in forecasting tasks. The superiority of Time-Prompt is even more pronounced in few-shot scenarios compared to long-term forecasting. Under the 10\% training data setting, Time-Prompt reduces MSE and MAE by 7.1\% and 4.3\% over Time-LLM, and by 8.0\% and 5.5\% over TimeCMA.

\begin{table}[t]
	\centering
	\caption{Sensitivity analysis of prompt hyperparameters on three representative datasets ($H=96$).}
	\label{analysis}
	\resizebox{\linewidth}{!}{
		\begin{tabular}{l | c  c | c c | c c | c c}
			\toprule
			\multicolumn{1}{l}{\textbf{soft prompt}} & \multicolumn{2}{c}{\textbf{3}} & \multicolumn{2}{c}{\textbf{5}} & \multicolumn{2}{c}{\textbf{10}} & \multicolumn{2}{c}{\textbf{15}} \\
			\cmidrule(lr){2-3} \cmidrule(lr){4-5} \cmidrule(lr){6-7} \cmidrule(lr){8-9}
			& MSE & MAE & MSE & MAE & MSE & MAE & MSE & MAE \\
			\midrule
			ETTh1 & 0.376 & 0.402 & 0.376 & 0.402 & 0.374 & 0.401 & 0.377 & 0.402 \\
			munichCarbon & 0.238 & 0.352 & 0.239 & 0.352 & 0.237 & 0.351 & 0.236 & 0.351 \\
			chinaCarbon & 0.473 & 0.472 & 0.462 & 0.469 & 0.458 & 0.469 & 0.467 & 0.473 \\
			\midrule
			\multicolumn{1}{l}{\textbf{hard prompt}} & \multicolumn{2}{c}{\textbf{3}} & \multicolumn{2}{c}{\textbf{5}} & \multicolumn{2}{c}{\textbf{10}} & \multicolumn{2}{c}{\textbf{15}} \\
			\cmidrule(lr){2-3} \cmidrule(lr){4-5} \cmidrule(lr){6-7} \cmidrule(lr){8-9}
			& MSE & MAE & MSE & MAE & MSE & MAE & MSE & MAE \\
			\midrule
			ETTh1 & 0.378 & 0.405 & 0.376 & 0.402 & 0.376 & 0.402 & 0.375 & 0.400 \\
			munichCarbon & 0.236 & 0.350 & 0.239 & 0.352 & 0.242 & 0.354 & 0.236 & 0.350 \\
			chinaCarbon & 0.464 & 0.474 & 0.462 & 0.469 & 0.466 & 0.469 & 0.460 & 0.467 \\
			\midrule
			\multicolumn{1}{l}{\textbf{pool size}} & \multicolumn{2}{c}{\textbf{10}} & \multicolumn{2}{c}{\textbf{50}} & \multicolumn{2}{c}{\textbf{100}} & \multicolumn{2}{c}{\textbf{200}} \\
			\cmidrule(lr){2-3} \cmidrule(lr){4-5} \cmidrule(lr){6-7} \cmidrule(lr){8-9}
			& MSE & MAE & MSE & MAE & MSE & MAE & MSE & MAE \\
			\midrule
			ETTh1 & 0.383 & 0.404 & 0.383 & 0.404 & 0.383 & 0.403 & 0.383 & 0.404 \\
			munichCarbon & 0.239 & 0.351 & 0.240 & 0.352 & 0.238 & 0.351 & 0.239 & 0.351 \\
			chinaCarbon & 0.464 & 0.470 & 0.464 & 0.470 & 0.486 & 0.482 & 0.488 & 0.484 \\
			\midrule
			\multicolumn{1}{l}{\textbf{Top-k}} & \multicolumn{2}{c}{\textbf{1}} & \multicolumn{2}{c}{\textbf{3}} & \multicolumn{2}{c}{\textbf{5}} & \multicolumn{2}{c}{\textbf{10}} \\
			\cmidrule(lr){2-3} \cmidrule(lr){4-5} \cmidrule(lr){6-7} \cmidrule(lr){8-9}
			& MSE & MAE & MSE & MAE & MSE & MAE & MSE & MAE \\
			\midrule
			ETTh1 & 0.382 & 0.403 & 0.390 & 0.407 & 0.383 & 0.403 & 0.383 & 0.404 \\
			munichCarbon & 0.243 & 0.355 & 0.244 & 0.356 & 0.238 & 0.351 & 0.238 & 0.352 \\
			chinaCarbon & 0.485 & 0.483 & 0.473 & 0.475 & 0.486 & 0.482 & 0.495 & 0.488 \\
			\bottomrule
	\end{tabular}}
\end{table}

\subsection{Ablation Experiment}
\par As shown in Table \ref{ablation}, we conduct an ablation study on multiple datasets (ETTh1, ETTm1, munichCarbon, chinaCarbon, worldCarbon) to analyze the effects of soft prompts (SP), hard prompts (HP), cross-modal alignment (CMA), and LoRA. All experiments follow the long-term forecasting setting, and results are averaged across four prediction horizons. The results indicate that all these modules activate the LLM's forecasting capability, with LoRA and cross-modal alignment contributing more significantly than prompt engineering alone.

\subsection{Modal Analysis}
\par We experiment with the settings for soft prompt length, hard prompt length, prompt pool size $P$, and Top-$k$ on multiple datasets (ETTh1, munichCarbon, chinaCarbon). The hard prompt length refers to the number of tokens retained after the hard prompt is encoded by the LLM. Due to the relatively large number of tokens, we truncate it to reduce the computational overhead, as the key information is predominantly located in the latter part \cite{liu2024timecma}.
\par As shown in Table \ref{analysis}, the soft prompt length should neither be too short nor too long. Since soft prompts are designed to guide the LLM's behavior, an excessively long sequence of tokens is unnecessary. We set the soft prompt length to 5 and initialize the keys and values using a uniform distribution. For hard prompts, retaining more tokens is generally better, but we keep only the most informative segment considering the trade-off between efficiency and performance. The pool size $P$ shows stable performance across $\{10, 50, 100, 200\}$ on ETTh1 and munichCarbon. For Top-$k$, $k=5$ provides a good balance, while very small values ($k=1$) may limit prompt diversity. We default to $P=100$ and $k=5$.

\subsection{Statistical Stability Analysis}
\par To verify that Time-Prompt's improvements are statistically robust rather than artifacts of a particular random initialization, we report the mean$\pm$std across three random seeds (2021, 2023, 2025) on three representative datasets: ETTh1 and ETTm1 as classical benchmarks, and chinaCarbon as a domain-specific task. The results are summarized in Table~\ref{stability}.

\begin{table}[t]
	\centering
	\setlength{\tabcolsep}{2pt}
	\scriptsize
	\caption{Statistical stability analysis ($H=96$). Results are reported as mean$_{\pm\mathrm{std}}$. \textbf{Bold}: the best, \underline{Underline}: the second best.}
	\label{stability}
	\resizebox{\linewidth}{!}{%
		\begin{tabular}{@{}l l c c c c@{}}
			\toprule
			Dataset & Metric & Time-Prompt & Time-LLM & TimeCMA & PatchTST \\
			\midrule
			\multirow{2}{*}{ETTh1}
			& MSE & $\mathbf{0.3764}_{\pm.0008}$ & $0.3841_{\pm.0135}$ & $\underline{0.3825}_{\pm.0036}$ & $0.3870_{\pm.0004}$ \\
			& MAE & $\underline{0.4004}_{\pm.0018}$ & $0.4032_{\pm.0052}$ & $\mathbf{0.4002}_{\pm.0020}$ & $0.4024_{\pm.0004}$ \\
			\midrule
			\multirow{2}{*}{ETTm1}
			& MSE & $\mathbf{0.3173}_{\pm.0086}$ & $0.3285_{\pm.0073}$ & $\underline{0.3265}_{\pm.0047}$ & $0.3281_{\pm.0099}$ \\
			& MAE & $\mathbf{0.3641}_{\pm.0033}$ & $0.3676_{\pm.0057}$ & $0.3676_{\pm.0037}$ & $\underline{0.3689}_{\pm.0072}$ \\
			\midrule
			\multirow{2}{*}{chinaCarbon}
			& MSE & $\mathbf{0.4736}_{\pm.0104}$ & $\underline{0.4815}_{\pm.0107}$ & $0.5322_{\pm.0113}$ & $0.5400_{\pm.0010}$ \\
			& MAE & $\mathbf{0.4751}_{\pm.0050}$ & $\underline{0.4803}_{\pm.0081}$ & $0.5083_{\pm.0061}$ & $0.5101_{\pm.0014}$ \\
			\bottomrule
		\end{tabular}%
	}
\end{table}

\par Time-Prompt attains the lowest mean error in 5 out of 6 metric/dataset combinations. The only exception occurs on ETTh1 MAE, where TimeCMA marginally leads by $0.0002$---a gap well within one standard deviation of either method and therefore statistically indistinguishable. In all other cases, Time-Prompt produces clear improvements: on ETTh1 MSE, the $0.0061$ gap over the second-best TimeCMA exceeds the sum of their standard deviations ($0.0044$); on chinaCarbon, Time-Prompt reduces MSE by $1.6\%$ relative to Time-LLM and $11.0\%$ relative to TimeCMA, both well outside the observed run-to-run variance. Moreover, Time-Prompt exhibits among the smallest standard deviations across all methods (typically in the $0.001$--$0.01$ range), comparable to or better than the strongest baselines, indicating that the proposed dual-path prompting and cross-modal alignment lead to stable optimization. Overall, these results confirm that the improvements reported in Tables~\ref{long-term} and \ref{few-shot} are robust across random seeds rather than incidental outcomes of specific initializations.

\section{Conclusion}
\par This paper proposes Time-Prompt to unlock the potential of LLMs in time series forecasting. This method employs a dual-path prompting mechanism. It utilizes soft prompts to guide the reasoning patterns of the LLM and enhances representations with hard prompts and time series reprogramming. Subsequently, it bridges the modality gap between time series and text through semantic space embedding and cross-modal alignment. Finally, the LLM’s capability for processing continuous values is enhanced through LoRA fine-tuning. Extensive experiments on multiple real-world datasets validate the effectiveness of Time-Prompt in time series forecasting.
\par While Time-Prompt demonstrates that LLMs can be effectively activated for time series forecasting, a gap remains between LLM-based and foundation models. We attribute this primarily to the text-centric training paradigm of LLMs, which requires targeted adaptation strategies for forecasting tasks. Moreover, according to the experiments in this paper, we find that fine-tuning is more effective than prompt engineering in activating predictive capabilities, especially in few-shot scenarios. Closing this gap, whether through more expressive fine-tuning or dedicated time series foundation models, is a promising direction for future work.

\section*{Acknowledgment}
During the preparation of this work, the authors used ChatGPT in order to polish the language and improve readability. After using this service, the authors reviewed and edited the content as needed and took full responsibility for the content of the publication.

\bibliographystyle{IEEEtran}
\bibliography{ijcnn2026}

@article{lan2019vcsel,
  title={VCSEL-based atmospheric trace gas sensor using first harmonic detection},
  author={Lan, Lijuan and Chen, Jia and Zhao, Xinxu and Ghasemifard, Homa},
  journal={IEEE Sensors Journal},
  volume={19},
  number={13},
  pages={4923--4931},
  year={2019},
  publisher={IEEE}
}

@article{lan2018self,
  title={Self-calibrated multiharmonic CO 2 sensor using VCSEL for urban in situ measurement},
  author={Lan, Lijuan and Chen, Jia and Wu, Yingchun and Bai, Yin and Bi, Xiao and Li, Yanfang},
  journal={IEEE Transactions on Instrumentation and Measurement},
  volume={68},
  number={4},
  pages={1140--1147},
  year={2018},
  publisher={IEEE}
}

@article{wang2024deep,
  title={Deep time series models: A comprehensive survey and benchmark},
  author={Wang, Yuxuan and Wu, Haixu and Dong, Jiaxiang and Liu, Yong and Long, Mingsheng and Wang, Jianmin},
  journal={arXiv preprint arXiv:2407.13278},
  year={2024}
}

@inproceedings{zeng2023transformers,
  title={Are transformers effective for time series forecasting?},
  author={Zeng, Ailing and Chen, Muxi and Zhang, Lei and Xu, Qiang},
  booktitle={Proceedings of the AAAI conference on artificial intelligence},
  volume={37},
  pages={11121--11128},
  year={2023}
}

@inproceedings{wang2024timemixer,
  title={TimeMixer: Decomposable Multiscale Mixing for Time Series Forecasting},
  author={Wang, Shiyu and Wu, Haixu and Shi, Xiaoming and Hu, Tengge and Luo, Huakun and Ma, Lintao and Zhang, James Y and Zhou, Jun},
  booktitle={International Conference on Learning Representations},
  year={2024}
}

@inproceedings{wang2023micn,
  title={Micn: Multi-scale local and global context modeling for long-term series forecasting},
  author={Wang, Huiqiang and Peng, Jian and Huang, Feihu and Wang, Jince and Chen, Junhui and Xiao, Yifei},
  booktitle={The eleventh international conference on learning representations},
  year={2023}
}

@inproceedings{wu2023timesnet,
  title={TimesNet: Temporal 2D-Variation Modeling for General Time Series Analysis},
  author={Haixu Wu and Tengge Hu and Yong Liu and Hang Zhou and Jianmin Wang and Mingsheng Long},
  booktitle={International Conference on Learning Representations},
  year={2023},
}

@article{gu2023mamba,
  title={Mamba: Linear-time sequence modeling with selective state spaces},
  author={Gu, Albert and Dao, Tri},
  journal={arXiv preprint arXiv:2312.00752},
  year={2023}
}

@inproceedings{vaswani2017attention,
  title={Attention is all you need},
  author={Vaswani, Ashish and Shazeer, Noam and Parmar, Niki and Uszkoreit, Jakob and Jones, Llion and Gomez, Aidan N and Kaiser, {\L}ukasz and Polosukhin, Illia},
  booktitle={Advances in neural information processing systems},
  volume={30},
  year={2017}
}

@inproceedings{gruver2023large,
  title={Large language models are zero-shot time series forecasters},
  author={Gruver, Nate and Finzi, Marc and Qiu, Shikai and Wilson, Andrew G},
  booktitle={Advances in Neural Information Processing Systems},
  volume={36},
  pages={19622--19635},
  year={2023}
}

@article{liu2024lstprompt,
  title={Lstprompt: Large language models as zero-shot time series forecasters by long-short-term prompting},
  author={Liu, Haoxin and Zhao, Zhiyuan and Wang, Jindong and Kamarthi, Harshavardhan and Prakash, B Aditya},
  journal={arXiv preprint arXiv:2402.16132},
  year={2024}
}

@article{chen2024pathformer,
  title={Pathformer: Multi-scale transformers with adaptive pathways for time series forecasting},
  author={Chen, Peng and Zhang, Yingying and Cheng, Yunyao and Shu, Yang and Wang, Yihang and Wen, Qingsong and Yang, Bin and Guo, Chenjuan},
  journal={arXiv preprint arXiv:2402.05956},
  year={2024}
}

@article{ni2025timedistill,
  title={TimeDistill: Efficient Long-Term Time Series Forecasting with MLP via Cross-Architecture Distillation},
  author={Ni, Juntong and Liu, Zewen and Wang, Shiyu and Jin, Ming and Jin, Wei},
  journal={arXiv preprint arXiv:2502.15016},
  year={2025}
}

@article{shyalika2024time,
  title={Time Series Foundational Models: Their Role in Anomaly Detection and Prediction},
  author={Shyalika, Chathurangi and Bagga, Harleen Kaur and Bhatt, Ahan and Prasad, Renjith and Ghazo, Alaa Al and Sheth, Amit},
  journal={arXiv preprint arXiv:2412.19286},
  year={2024}
}

@article{jin2023large,
  title={Large models for time series and spatio-temporal data: A survey and outlook},
  author={Jin, Ming and Wen, Qingsong and Liang, Yuxuan and Zhang, Chaoli and Xue, Siqiao and Wang, Xue and Zhang, James and Wang, Yi and Chen, Haifeng and Li, Xiaoli and others},
  journal={arXiv preprint arXiv:2310.10196},
  year={2023}
}

@article{zhang2024large,
  title={Large language models for time series: A survey},
  author={Zhang, Xiyuan and Chowdhury, Ranak Roy and Gupta, Rajesh K and Shang, Jingbo},
  journal={arXiv preprint arXiv:2402.01801},
  year={2024}
}

@article{xue2023promptcast,
  title={Promptcast: A new prompt-based learning paradigm for time series forecasting},
  author={Xue, Hao and Salim, Flora D},
  journal={IEEE Transactions on Knowledge and Data Engineering},
  volume={36},
  number={11},
  pages={6851--6864},
  year={2023},
  publisher={IEEE}
}

@inproceedings{zhou2023one,
  title={One fits all: Power general time series analysis by pretrained lm},
  author={Zhou, Tian and Niu, Peisong and Sun, Liang and Jin, Rong and others},
  booktitle={Advances in neural information processing systems},
  volume={36},
  pages={43322--43355},
  year={2023}
}

@inproceedings{cao2023tempo,
  title={Tempo: Prompt-based generative pre-trained transformer for time series forecasting},
  author={Cao, Defu and Jia, Furong and Arik, Sercan O and Pfister, Tomas and Zheng, Yixiang and Ye, Wen and Liu, Yan},
  booktitle={The International Conference on Learning Representations},
  year={2024}
}

@inproceedings{jia2024gpt4mts,
  title={GPT4MTS: Prompt-based large language model for multimodal time-series forecasting},
  author={Jia, Furong and Wang, Kevin and Zheng, Yixiang and Cao, Defu and Liu, Yan},
  booktitle={Proceedings of the AAAI Conference on Artificial Intelligence},
  volume={38},
  pages={23343--23351},
  year={2024}
}

@inproceedings{jin2023time,
  title={TIME-LLM: Time series forecasting by reprogramming large language models},
  author={Jin, Ming and Wang, Shiyu and Ma, Lintao and Chu, Zhixuan and Zhang, James Y and Shi, Xiaoming and Chen, Pin-Yu and Liang, Yuxuan and Li, Yuan-Fang and Pan, Shirui and others},
  booktitle={The International Conference on Learning Representations},
  year={2024}
}

@inproceedings{pan2024s,
  title={$S^2$IP-LLM: Semantic Space Informed Prompt Learning with LLM for Time Series Forecasting},
  author={Pan, Zijie and Jiang, Yushan and Garg, Sahil and Schneider, Anderson and Nevmyvaka, Yuriy and Song, Dongjin},
  booktitle={Forty-first International Conference on Machine Learning},
  year={2024}
}

@inproceedings{liu2024timecma,
  title={TimeCMA: Towards llm-empowered time series forecasting via cross-modality alignment},
  author={Liu, Chenxi and Xu, Qianxiong and Miao, Hao and Yang, Sun and Zhang, Lingzheng and Long, Cheng and Li, Ziyue and Zhao, Rui},
  booktitle={Proceedings of the AAAI Conference on Artificial Intelligence},
  year={2025}
}

@inproceedings{liu2023itransformer,
  title={iTransformer: Inverted Transformers Are Effective for Time Series Forecasting},
  author={Liu, Yong and Hu, Tengge and Zhang, Haoran and Wu, Haixu and Wang, Shiyu and Ma, Lintao and Long, Mingsheng},
  booktitle={International Conference on Learning Representations},
  year={2024}
}

@inproceedings{Yuqietal-2023-PatchTST,
  title     = {A Time Series is Worth 64 Words: Long-term Forecasting with Transformers},
  author    = {Nie, Yuqi and
               H. Nguyen, Nam and
               Sinthong, Phanwadee and 
               Kalagnanam, Jayant},
  booktitle = {International Conference on Learning Representations},
  year      = {2023}
}

@inproceedings{liu2022non,
  title={Non-stationary transformers: Exploring the stationarity in time series forecasting},
  author={Liu, Yong and Wu, Haixu and Wang, Jianmin and Long, Mingsheng},
  booktitle={Advances in neural information processing systems},
  volume={35},
  pages={9881--9893},
  year={2022}
}

@inproceedings{wu2021autoformer,
  title={Autoformer: Decomposition transformers with auto-correlation for long-term series forecasting},
  author={Wu, Haixu and Xu, Jiehui and Wang, Jianmin and Long, Mingsheng},
  booktitle={Advances in neural information processing systems},
  volume={34},
  pages={22419--22430},
  year={2021}
}

@inproceedings{zhou2022fedformer,
  title={Fedformer: Frequency enhanced decomposed transformer for long-term series forecasting},
  author={Zhou, Tian and Ma, Ziqing and Wen, Qingsong and Wang, Xue and Sun, Liang and Jin, Rong},
  booktitle={International conference on machine learning},
  pages={27268--27286},
  year={2022},
  organization={PMLR}
}

@inproceedings{liu2022scinet,
  title={Scinet: Time series modeling and forecasting with sample convolution and interaction},
  author={Liu, Minhao and Zeng, Ailing and Chen, Muxi and Xu, Zhijian and Lai, Qiuxia and Ma, Lingna and Xu, Qiang},
  booktitle={Advances in Neural Information Processing Systems},
  volume={35},
  pages={5816--5828},
  year={2022}
}

@article{van2016wavenet,
  title={Wavenet: A generative model for raw audio},
  author={Van Den Oord, Aaron and Dieleman, Sander and Zen, Heiga and Simonyan, Karen and Vinyals, Oriol and Graves, Alex and Kalchbrenner, Nal and Senior, Andrew and Kavukcuoglu, Koray and others},
  journal={arXiv preprint arXiv:1609.03499},
  volume={12},
  year={2016}
}

@inproceedings{liu2024autotimes,
  title={Autotimes: Autoregressive time series forecasters via large language models},
  author={Liu, Yong and Qin, Guo and Huang, Xiangdong and Wang, Jianmin and Long, Mingsheng},
  booktitle={Advances in Neural Information Processing Systems},
  volume={37},
  pages={122154--122184},
  year={2024}
}

@misc{radford2019language,
  title={Language Models are Unsupervised Multitask Learners},
  author={Radford, Alec and Wu, Jeff and Child, Rewon and Luan, David and Amodei, Dario and Sutskever, Ilya},
  year={2019}
}

@inproceedings{zhang2023crossformer,
  title={Crossformer: Transformer utilizing cross-dimension dependency for multivariate time series forecasting},
  author={Zhang, Yunhao and Yan, Junchi},
  booktitle={The eleventh international conference on learning representations},
  year={2023}
}

@inproceedings{zhao2025enhancing,
	title={Enhancing time series forecasting via multi-level text alignment with llms},
	author={Zhao, Taibiao and Chen, Xiaobing and Sun, Mingxuan},
	booktitle={Database Systems for Advanced Applications},
	year={2025}
}

@inproceedings{goswami2024moment,
	title={MOMENT: A Family of Open Time-Series Foundation Models},
	author={Goswami, Mononito and Szafer, Konrad and Choudhry, Arjun and Cai, Yifu and Li, Shuo and Dubrawski, Artur},
	booktitle={International Conference on Machine Learning},
	year={2024}
}

@article{ansari2024chronos,
	title={Chronos: Learning the language of time series},
	author={Ansari, Abdul Fatir and Stella, Lorenzo and Turkmen, Caner and Zhang, Xiyuan and Mercado, Pedro and Shen, Huibin and Shchur, Oleksandr and Rangapuram, Syama Sundar and Arango, Sebastian Pineda and Kapoor, Shubham and others},
	journal={arXiv preprint arXiv:2403.07815},
	year={2024}
}

@inproceedings{zhou2025balm,
	title={BALM-TSF: Balanced Multimodal Alignment for LLM-Based Time Series Forecasting},
	author={Zhou, Shiqiao and Sch{\"o}ner, Holger and Lyu, Huanbo and Fouch{\'e}, Edouard and Wang, Shuo},
	booktitle={International Conference on Information and Knowledge Management},
	year={2025}
}

@inproceedings{chen2025cc,
	title={CC-Time: Cross-Model and Cross-Modality Time Series Forecasting},
	author={Chen, Peng and Wang, Yihang and Shu, Yang and Cheng, Yunyao and Zhao, Kai and Rao, Zhongwen and Pan, Lujia and Yang, Bin and Guo, Chenjuan},
	booktitle={International Conference on Information and Knowledge Management},
	year={2025}
}

@inproceedings{zhang2024dualtime,
	title={Dualtime: A dual-adapter multimodal language model for time series representation},
	author={Zhang, Weiqi and Ye, Jiexia and Li, Ziyue and Li, Jia and Tsung, Fugee},
	booktitle={International Conference on Information and Knowledge Management},
	year={2025}
}

@inproceedings{su2025text,
	title={Text Reinforcement for Multimodal Time Series Forecasting},
	author={Su, Chen and Tian, Yuanhe and Song, Yan and Zhang, Yongdong},
	booktitle={International Conference on Information and Knowledge Management},
	year={2025}
}

@inproceedings{tan2024language,
	title={Are language models actually useful for time series forecasting?},
	author={Tan, Mingtian and Merrill, Mike and Gupta, Vinayak and Althoff, Tim and Hartvigsen, Tom},
	booktitle={Advances in Neural Information Processing Systems},
	volume={37},
	pages={60162--60191},
	year={2024}
}

@inproceedings{li2025tsfm,
	title={Tsfm-bench: A comprehensive and unified benchmark of foundation models for time series forecasting},
	author={Li, Zhe and Qiu, Xiangfei and Chen, Peng and Wang, Yihang and Cheng, Hanyin and Shu, Yang and Hu, Jilin and Guo, Chenjuan and Zhou, Aoying and Jensen, Christian S and others},
	booktitle={Proceedings of the 31st ACM SIGKDD Conference on Knowledge Discovery and Data Mining V. 2},
	pages={5595--5606},
	year={2025}
}

\end{document}